\documentclass{article}
\usepackage{spconf,amsmath,epsfig}
\usepackage{graphicx}
\usepackage{multirow}
\usepackage{verbatim}
\usepackage{float}
\usepackage{array}
\usepackage{amssymb}
\usepackage[tight,footnotesize]{subfigure}
\usepackage[colorlinks,
linkcolor=red,
anchorcolor=blue,
citecolor=green
]{hyperref}
\hypersetup{bookmarksdepth=-3}

\newcommand\blfootnote[1]{%
	\begingroup
	\renewcommand\thefootnote{}\footnote{#1}%
	\addtocounter{footnote}{-1}%
	\endgroup
}
\usepackage{fancyhdr}
\begin{document}\sloppy
	
	% Example definitions.
	% --------------------
	\def\x{{\mathbf x}}
	\def\L{{\cal L}}

	% Title.
	% ------
	\title{JUSTLOOKUP:\ One Millisecond Deep Feature Extraction \\
		for Point Clouds by Lookup Tables}
	%
	% Single address.
	% ---------------
	\name{Hongxin Lin$^1$$^,$$^2$$^,$$^\dagger$,
		Zelin Xiao$^1$$^,$$^2$$^,$$^\dagger$,
		Yang Tan$^1$$^,$$^2$,
		Hongyang Chao$^1$,
		Shengyong Ding$^1$$^,$$^*$}
	\address{$^{1}$School of Data and Computer Science, Sun Yat-sen University, Guangzhou, China \\
	 $^{2}$Pixtalks Tech, Guangzhou, China \\
		\{linhx9, xiaozl, tany36\}@mail2.sysu.edu.cn, isschhy@mail.sysu.edu.cn, marcding@163.com}
	\begin{comment}			
	\name{Anonymous ICME submission}
	\address{}	
	\end{comment}		
	\maketitle

	\begin{abstract}
		Deep models are capable of fitting complex high dimensional functions while usually yielding large computation load. There is no way to speed up the inference process by classical lookup tables due to the high-dimensional input and limited memory size. Recently, a novel architecture (PointNet) for point clouds has demonstrated that it is possible to obtain a complicated deep function from a set of 3-variable functions. In this paper, we exploit this property and apply a lookup table to encode these 3-variable functions. This method ensures that the inference time is only determined by the memory access no matter how complicated the deep function is. We conduct extensive experiments on ModelNet and ShapeNet datasets and demonstrate that we can complete the inference process in 1.5 ms on an Intel i7-8700 CPU (single core mode), $32\times$ speedup over the  PointNet architecture without any performance degradation.
		
	\end{abstract}
	\blfootnote{$\dagger$ indicates equal contributions and * indicates the corresponding author.}
	\begin{keywords}
		point cloud, lookup table, 3D object classification, 3D object retrieval
	\end{keywords}

	\section{Introduction}	
	Learning representations directly from 3D point clouds is very attractive considering the intrinsic advantages of being less sensitive to pose and light changes. Deep neural models have demonstrated powerful ability of learning complicated functions on point cloud tasks \cite{qi2017pointnet, qi2017pointnet++, wang2018dynamic, li2018so}. However, such ability often comes at the cost of high computational demand which is prohibitive for applying deep 3D models on a wide range of devices. Therefore it is very desirable to reduce the computation complexity. Typical approaches include quantization or simplification of network architectures \cite{han2015deep, denton2014exploiting, ba2014deep}. Nevertheless, such approaches are still complicated and not fast enough for devices with limited computing capability.

	Apart from the recently proposed speedup methods for deep models, lookup table is a more classical and faster means to approximate complicated function values. This technique essentially divides the input space into non-overlapping regions and uses one single value to represent the function value for each region. Unfortunately, it is not applicable for general deep models due to the high-dimensional input and limited memory size. 

	\begin{comment}
			To see this, suppose we need to process a point cloud of size 1,000, the network input would be 3,000 dimensional and we need to allocate $O(2^{3,000})$ memory space to store the approximated values even we split each dimension into two intervals.
	\end{comment}
	\begin{comment}
		Our Sampled PointWise is refreshingly simple and fast.
		We firstly define a point-wise deep model, which is composed
		of a multi-layer perception(MLP) module and a max-pooling
		module. Our architecture is surprisingly simple as each point
		is processed identically and independently by MLP, which we
		define as a set of 3-variable deep function. Meanwhile, each
		point is represented by just its three coordinates and scaled to
		a fix sized volume. These special characteristics make it pos-
		sible to replace the network inference with classical lookup
		table, which can dramatically speed up the inference process.
		The crucial step of our approach is to convert the 3-
		variable deep function set into a lookup table, i.e. apply
		voxelization on the input space and use one single value to
		encode the actual function values on each voxel. Hence we
		can directly use the lookup table to retrieve the approximated
		function values from memory rather than network inference
		process
	\end{comment}
	
	Recently, Qi \textit{et al.} \cite{qi2017pointnet} proposed a deep architecture, i.e. PointNet \cite{qi2017pointnet} which showed that it is possible to learn a
	set of optimization functions and use max pooling operation to select the informative points of the point cloud. The basic architecture of PointNet \cite{qi2017pointnet} is extremely simple as each point is just represented by its three coordinates and each point is processed identically and independently by a multi-layer perception, which can be viewed as a set of 3-variable functions from a mathematical perspective.
	
	Inspired by this work, we propose to apply the lookup table to approximate  these 3-variable functions. More precisely, we first train a point-wise architecture to obtain a set of 3-variable functions. After that, we subdivide the fixed-size input space into equally spaced voxels and use one single value to encode the actual function value for each voxel.  Hence we
	can directly apply the lookup table to retrieve the approximate
	function value from memory rather than network inference.
	
	\begin{comment}
		Due to the 3-dimensional input and point-wise property, we can use a lookup table to approximate the deep function $h$, i.e. apply voxelization on the input space and use one single value to encode the actual function values on each voxel. After that we can directly use the lookup table to retrieve the approximated function value from a memory address rather than network inference process.

	More precisely, for a well optimized point wise network, we first design and train a point-wise architecture to obtain the aforescale the input to a fixed size volume and voxelize the volume into a set of voxels. For each voxel, we use one single value to encode the actual function values. Hence we can directly use the lookup table to retrieve the approximated function values from memory rather than network inference process. 
	\end{comment}
	Furthermore, we fine-tune model to adapt to the approximate function value so as to avoid the performance drop. We conduct extensive experiments on different  benchmark datasets. Experiments show that we can complete the inference process in 1.5 ms using only 15MB memory, 32$\times$ speedup over PointNet \cite{qi2017pointnet} while maintaining almost the same performance.

	\begin{comment}
	One concern with our architecture is that it lacks of the ability of capturing local structures. We argue that our network can implicitly learn local structures to some extent which is automatically achieved by max operation. This is derived from the following fact. At the critical point where each sketching function $h_s$ achieves its maximum value $T$, the level set of $h_s$, i.e. $\{x:h_s(x)=T\}$ is tangent to the object shape. This tangent constraints can implicitly encode the local structure. In addiction, when the number of sketching function $h_s$ is large enough, the level sets of  $h$ themselves can form neighboring relationship spatially, thus carrying the local structure information
	\end{comment}	
	One concern with the point-wise architecture is that it lacks of the ability of capturing local structures. We argue that the point-wise architecture can implicitly learn local structures to some extent, which is automatically achieved by max pooling operation. More interestingly, it turns out that the point-wise architecture can learn to enable the level sets (isosurfaces) of learned functions to be tangent to the object shape owing to the max pooling operation, thus carrying the local structure information.

	In summary, the \textbf{key contributions} of this paper are as follows:
	
	\begin{itemize}

		\item A method that can dramatically speed up the inference process with lookup table techniques for the point-wise deep architecture.
		
		\item A novel geometric interpretation for the point-wise deep architecture which can help understand the effectiveness of this architecture.
		
		\item Extensive experiments that demonstrate the correctness and effectiveness of our method.
	\end{itemize}
	\section{Related Work}
	\subsection{Deep Learning on 3D Point Cloud}
	PointNet \cite{qi2017pointnet} is the pioneer to apply deep neural networks to directly process unordered point clouds. To address the problem, it adopted spatial transform networks and a symmetry function to maintain the invariance of permutation. After that, many recent
	works mainly focus on how to efficiently capture local features based on PointNet \cite{qi2017pointnet}.
	For instance, PointNet++ \cite{qi2017pointnet++} applied PointNet \cite{qi2017pointnet} structure in local point sets with different resolutions and accumulated local features in a hierarchical architecture. In DGCNN \cite{wang2018dynamic}, EdgeConv was proposed as a basic block to build
	networks, in which the edge features between points and their neighbors were exploited. SO-Net \cite{li2018so}  explicitly modeled the spatial distribution of input points and systematically adjusted the receptive field overlap to perform hierarchical feature extraction.
	
	\subsection{Deep Model Compression and Acceleration}
	During the past few years, tremendous progress has been made in how to perform model compression and acceleration in deep networks without significantly degrading the model performance. To begin with, Han \textit{et al.} \cite{han2015deep} first pruned the unimportant connections and retrained the sparsely connected networks. Then they quantized the linked weights using weight sharing and applied Huffman coding to the quantized weights. Moreover, low-rank approximation and clustering schemes for the convolutional kernels were proposed in \cite{denton2014exploiting}. They achieved $2\times$ speedup for a single convolutional layer with 1\% drop on classification
	accuracy. Another well known method is Knowledge Distillation (KD), which has been recently adopted in \cite{ba2014deep} to compress deep and wide networks into shallower ones, where the compressed model mimics the function learned by the complex model.
	
	In this paper, we adopt an extremely simple strategy to dramatically speed up the inference process  by a lookup table for a particular network architecture, i.e. point-wise  architecture, without complicated pruning and quantization.
	\begin{figure}[t] \centering \small
		\includegraphics[width=0.5\textwidth,height = 0.18\textheight]{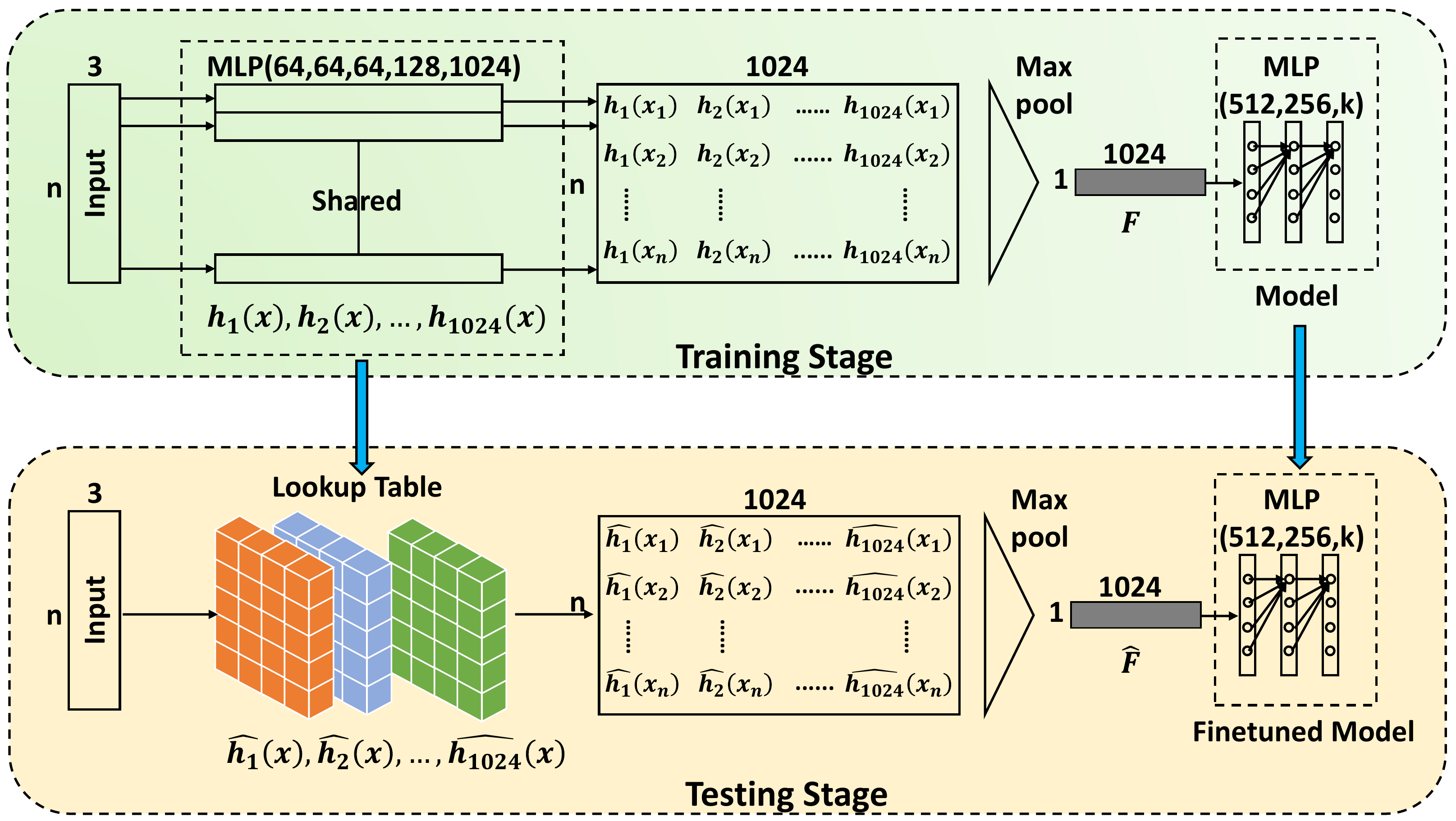}
		\caption{Overview of our method. During the training stage,  we use a multi-layer perception (MLP) module to implement each 3-variable function $h_s$ $(1\leq s \leq 1024)$ and then apply max pooling to obtain $\mathcal{F}$. For a specific point cloud task, we subsequently stack the model $M$ on $\mathcal{F}$ with a task-oriented loss function and jointly optimize the end-to-end network to obtain $\mathcal{F}$. Once we get $\mathcal{F}$, we can approximate all $h_s$ by a lookup table and then quickly get the approximate function $\hat{\mathcal{F}}$ during the testing stage. Furthermore, we fine-tune the model $M$ to adapt to the approximate  function $\hat{\mathcal{F}}$.}\label{tag:fig_architecture}
	\end{figure}
	\section{Proposed Method} \label{method}

	The main idea of our method is to train a point-wise architecture  for point cloud to get the point-wise function $\mathcal{F}$ and  then use lookup table techniques  to  quickly obtain the approximate function  $\hat{\mathcal{F}}$  with a simple array indexing operation and max operation.
	\begin{comment}
	
	Let us denote the function of a deep model by $y = \mathcal{F}(\mathcal{X})$ where $\mathcal{X}$  is the input and $y$ is the output. Our method is essentially to convert $\mathcal{F}$ into a lookup table and quickly obtain the approximate function value  $\hat y$  by memory access  with a simple array indexing operation rather than network inference.		
	Considering the limited memory capacity, we cannot apply this method on general deep functions as the input $\mathcal X$ is usually high-dimensional and the size of lookup table grows exponentially with the increase of dimensionality.
	Recently, a particular deep architecture for point cloud tasks, i.e. PointNet\cite{qi2017pointnet} has shown that it is possible to construct a powerful deep function from a set of point-wise functions(input variable is 3 dimensional).
	This offers us the possibility to design a point-wise architecture for point cloud so that  we can use a lookup table to extremely speed up the inference time called as Sampled PointWise.
	\end{comment}
	
	Formally, let $\mathcal{X} = \{x_p:x_p \in [-1,1]^3 \ \text{,} \ 1 \leq p \leq n \}$ represents the point cloud and each $x_p$ encodes the coordinates of a single point. We requires our $\mathcal{F}(\mathcal X)$ as following :
	\begin{equation}
	\mathcal{F}(\mathcal X)=(\max\{h_1(x_p)\}\text{,} \ \max\{h_2(x_p)\}\text{,}  ...\text{,}\ \max\{h_m(x_p)\})
	\end{equation}
	\begin{comment}
	in which $\mathcal{H}$ is defined as: $\{h_s: \mathbb{R}^3 \rightarrow \mathbb{R}\ \text{,} \ 1 \leq s \leq m\}$, i.e. a set of 3-variable functions. The $\mathcal{H}$ is implemented by a shared multi-layer perception  with 5 layers (layer output sizes 64, 64, 64, 128, 1024) on each point.
	\end{comment}
	
	We use $\mathcal{H}$  to denote the set of 3-variable functions, i.e. $\{h_s: \mathbb{R}^3 \rightarrow \mathbb{R}\ \text{,} \ 1 \leq s \leq m\}$. The $\mathcal{H}$ is implemented by a shared multi-layer perception  of 5 layers with layer output sizes 64, 64, 64, 128, 1024 respectively. Note that the value of $h_s(x_p)$ is defined as the $s^{\text{th}}$ output of the final layer.

	In order to learn $\mathcal{F}$, we further apply another model $M$ on $\mathcal{F}(\mathcal{X})$ with a task-oriented loss function, e.g. softmax loss or triplet loss and jointly optimize the end-to-end network to obtain $\mathcal{F}$. Once $\mathcal{F}$ has been learned, we subsequently construct a lookup table to approximate $\mathcal{H}$ so as to obtain the approximate function $\hat{\mathcal{F}}$. 
	
	It is obvious to see that the crucial step is to construct  a lookup table for $\mathcal{H}$. For convenience, we assume the input is scaled to a fixed-size volume $V$ where $V = [-1,1]^3$  and subdivide $V$ into $S^3$ equally spaced voxels with voxel length $\delta = 2/S$. Therefore the input volume $V$ is divided as:
	\begin{equation}
	V  = \bigcup_{i,j,k}[i\delta,(i+1)\delta]\times[j\delta,(j+1)\delta]\times [k\delta,(k+1)\delta]
	\end{equation}
	where $(i,j,k) \in [0,S]^3$.
	
	We further construct a lookup table \ $T[i][j][k][s]$ \ for $\mathcal{H}$ as follows:
	\begin{equation}
	T[i][j][k][s] = h_s(i \delta,j \delta,k \delta), 1\leq\ s \leq m
	\end{equation}
	
	Through this lookup table, given a point \ $(u,v,w)$, we first calculate its index \ $(i,j,k)$ \ as :
	\begin{equation}
	(i,j,k) = (\lfloor \frac{u + 1}{\delta}\rfloor,\lfloor \frac{v + 1}{\delta}\rfloor,\lfloor \frac{w + 1}{\delta}\rfloor)
	\end{equation}
	
	Then we can use the value $T[i][j][k][s]$ as the approximate value $ \hat{h_s}(u,v,w)$. Therefore the calculation of $\hat{\mathcal{H}}$ will be extremely fast which can be only determined by memory access time no matter how complex it is. By doing this for $\hat{\mathcal{H}}$, we can quickly obtain a final approximate function $\hat{\mathcal{F}}$ by simple max operation on $\hat{\mathcal{H}}$.
	
	Obviously, we might suffer from performance drop if we directly feed $\hat{\mathcal{F}}$ to the trained model $M$ as $\hat{\mathcal{F}}$ does not exactly equal $\mathcal{F}$ while $M$ is optimized for $\mathcal{F}$. So in order to keep the performance, we need to fine-tune $M$ based on $\hat{\mathcal{F}}$ after we construct a look table. This operation dramatically improves the performance and we can even achieve comparable performance with $S=25$. Fig.~\ref{tag:fig_architecture} shows how $\hat{\mathcal{F}}$ is implemented by our method.
	
	It is evident that in order to construct a lookup table for $\mathcal{H}$, the overall memory demand will be $ mLS^3$ \ bits where $L$ represents the bits of the element used to store the function value.

	Therefore we can efficiently reduce the memory demand by quantizing the function value, i.e. decrease $L$. Suppose the minimum and maximum values of $h_s$ in the input volume are $MIN$ and $MAX$ respectively. Then we quantize an original real value $r$ to a quantized level $q$ as follows:
	\begin{align}
	q =\lfloor \frac{r-MIN}{MAX-MIN}*2^L \rfloor
	\end{align}
	
	Conversely given a quantized level $q$, we obtain its approximate value $\hat{r}$ by:
	\begin{equation}
	\hat{r} = \frac{q*(MAX - MIN)}{2^L} + MIN
	\end{equation}
	
	In our experiments, we have found that our method can still achieve comparable performance with $L=8$.

	It's worth mentioning that apart from the point-wise architecture, PointNet \cite{qi2017pointnet} also designs two T-Net modules (input transform and feature transform) to improve the model capability. The T-Net parameters are actually learned from the max-pooled features, which pays more attention to the global information and loses the point-wise property. Hence we cannot apply the lookup table techniques to speed up the inference process once we introduce the T-Net modules.
	
	\section{Geometric Interpretation}
	
		\begin{figure}[t] \centering \small
		\includegraphics[width=0.35\textwidth,height = 0.2\textheight]{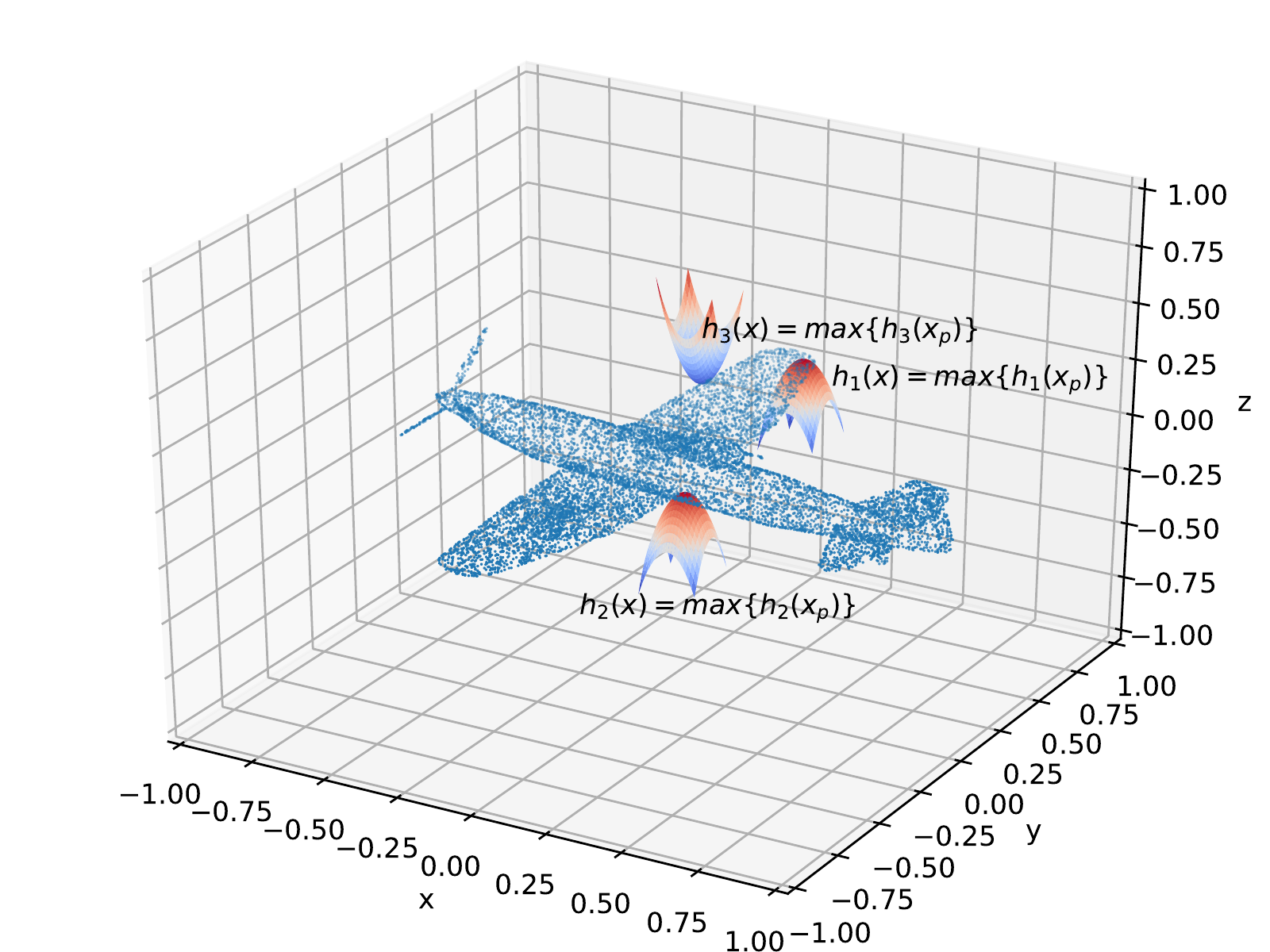}
		\caption{Illustration of how the level set (isosurface) of $h_s(x)$ $(1\leq s \leq m)$ is tangent to the shape which captures the local information surrounding the critical point.}\label{tag:tagent}
	\end{figure}
	One concern with the point-wise architecture is that it lacks of the ability of capturing the local structure as the key processing step is point-wise without using any neighborhood information explicitly. However, it still gains impressive results on 3D vision tasks.
	
	We argue that the point-wise network can indeed implicitly learn local structures. We explain this from a geometrical perspective. Use the aforementioned $h_s(x)$ $(1\leq s \leq m)$ to denote the 3-variable function. Evidently, the level set of $h_s(x)$, i.e. the set of points whose values are equal to a fixed value is a level set (an isosurface) in 3D space. The max pooling operation for each function $h_s$ can be considered as growing the level set until it touches the object shape at the critical point. It is not hard to see that at this critical point, the level set is tangent to the shape, thus capturing the local information surrounding the critical point to some extent. Furthermore, when the number
	of $h_s(x)$ is large enough, the collection of nearby level sets can
	also encode the local structure. Fig.~\ref{tag:tagent} shows how the level set of $h_s(x)$ captures the local structure of point cloud.
	\begin{table*}[t]
		\small
		\centering
		\caption{3D object classification results on ModelNet. The
			inference time for GPU is acquired with a batch size of 8 on a NVIDIA
			GTX1080 and the inference time for CPU is tested on an Intel i7-8700 CPU (single core mode) with a batch size of 1. }
		\vspace{6pt}
		{
			\setlength\tabcolsep{6pt} % default value: 6pt
			\begin{tabular}{ll|c|cc|cc|cc}
				\hline
				\multirow{2}{*}{Method} & \multirow{2}{*}{Representation} & \multirow{2}{*}{Input} & \multicolumn{2}{c|}{Inference Time} & \multicolumn{2}{c|}{ModelNet10} & \multicolumn{2}{c}{ModelNet40} \\
				&     &     & GPU (ms)      & CPU (ms)      & avg.class         & overall        & avg.class & overall       \\
				\hline
				PointNet++, \cite{qi2017pointnet++}& points + normal                          & $5000\times 6$                   & 163.2             & -               & -   & -                &-       & 91.9         \\
				DGCNN, \cite{wang2018dynamic}& points                          & $1024\times 3$                   & 94.6            & -               &-  & -              & 90.2      & 92.2         \\
				SO-Net, \cite{li2018so}& points + normal                          & $5000\times 6$                   & 59.6             & -               & \textbf{95.5}  & \textbf{95.7}               & \textbf{90.8}      & \textbf{93.4}         \\	
				PointNet, \cite{qi2017pointnet}& points                          & $1024\times 3$                   & 25.3             & 49.15               & -  & -               & 86.2      & 89.2         \\
				
				\hline
				PointWise         & points                  & $1024\times 3$                  & -             & 17  & 91.67    & 91.96                 & 84.98   & 87.95      \\
				\hline
				Ours (Sampled PointWise)          & points                  & $1024\times 3$                  & -             & \textbf{1.5} & 91.52     & 91.80                  & 84.89    & 87.84     \\
				\hline
				Ours (Sampled PointWise$_f$)          & points                  & $1024\times 3$                  & -             & \textbf{1.5}  & 92.08   & 92.85        & 86.43     & 89.51               \\
				\hline
			\end{tabular}
		\vspace*{-10pt}
		}
		
		\label{tbl_cls_table}
	\end{table*}
	
	\begin{table}[t]
		\centering
		\small
			\caption{Retrieval results on ModelNet10 and ModelNet40. The metric is tested in terms of mAP.}
			\vspace{6pt}
		\begin{tabular}{c|cc}
			\hline
			\multirow{2}{*}{Method}            & \multicolumn{2}{c}{mAP (\%)} \\
			&  ModelNet10  & ModelNet40 \\
			\hline
					PANORAMA-NN, \cite{sfikas2017exploiting}    & 87.40    & 83.50 \\
						DeepPano, \cite{shi2015deeppano}    & 84.18   & 76.81  \\
			PVNet, \cite{you2018pvnet}    & -   & \textbf{89.50}  \\
			SeqViews2SeqLabels, \cite{han2019seqviews2seqlabels}     & 91.43   & 89.09  \\
			PANORAMA-ENN, \cite{sfikas2018ensemble}   & \textbf{93.28}   & 86.34  \\
			GIFT, \cite{bai2016gift}   & 91.12   & 81.94  \\
			\hline
			PointWise  & 88.15   &  83.51    \\
			\hline
			Ours (Sampled PointWise)   & 87.98   & 83.09 \\
			\hline
			Ours (Sampled PointWise$_f$)    & 90.01   & 85.22 \\
			\hline
		\end{tabular}
	
		\label{tbl_modelnet_retrieval_table}
		\vspace{-6pt}
	\end{table}

	\section{Experiment}
	
	In this section, we first describe implementation details and datasets used for training and testing our method. Next, we compare our
	method with  a number of state-of-the-art methods on different benchmark datasets for 3D object classification and retrieval tasks.
	Finally, we provide detailed experiments under different numbers of voxels and further analyse our method's performance as we change the number of input points.
	\subsection{Implementation Detail}
	We first train a point-wise network to obtain a set of 3-variable functions for point cloud normalized to a fixed-size volume V where $V = [-1,1]^3$. For simplicity, we call the point-wise network as PointWise. Next, we set $S = 200$ and subdivide $V$ into $200\times 200 \times 200$ equally spaced voxels and apply a lookup table to encode the 3-variable functions we have obtained. Hence, during the testing phase we only need to retrieve function values from a lookup table instead of network inference, which we denote as Sampled PointWise. In the end, in order to avoid performance drop we fine-tune the model $M$ to adapt to the approximate function values, which we call Sampled PointWise$_f$.
	\subsection{DataSets for Training and Testing}
	Two variants of the ModelNet \cite{wu20153d}, i.e. ModelNet10 and
	ModelNet40, are used as the benchmarks for the classification
	task in Sec.~\ref{classification} and the retrieval task in Sec.~\ref{retrieval}.
	The ModelNet40 contains 12,311 models from 40  object categories, and the ModelNet10 contains 4,899 models from 10  object categories. In our experiment, we split ModelNet40 into 9,843 for training and 2,468 for testing and split ModelNet10 into 3,991 for training and 908 for testing.
	
	We also conduct experiments on ShapeNet-Core55 \cite{savva2016shrec} for the retrieval task in Sec.~\ref{retrieval}. ShapeNet-Core55 \cite{savva2016shrec} benchmark has two evaluation datasets: normal and perturbed. For normal dataset, all model data is
	consistently aligned while in perturbed dataset each model has been randomly rotated by a uniformly sampled rotation. In this paper, we only consider normal dataset which contains a total of 51,190 3D models with 55 categories. 70\% of the dataset is used for training, 10\% for validation, and 20\% for testing.

	\subsection{3D Object Classification} \label{classification}
	For 3D object classification tasks, we stack classification model $M$ (a multi-layer perception  with 3 layers) on $\mathcal{F}$ with softmax loss function. We evaluate our method on the ModelNet10 and ModelNet40  shape classification benchmarks. 
	
	In Table \ref{tbl_cls_table}  we show our Sampled PointWise$_f$ can gain a strong lead in speed while achieving comparable performance among methods based on 3D point cloud. Our inference time only needs 1.5ms by using 1024 points, $32\times$ speedup over PointNet \cite{qi2017pointnet} on an Intel i7-8700 CPU (single core mode) while
	maintaining the same performance. Indeed, we can obtain   a final $\hat{\mathcal{F}}$ through our lookup table in 0.9 ms while the time for the model $M$ (classification module) is 0.6 ms. Note that the basic architecture of our method is derived from PointNet \cite{qi2017pointnet} and even more simple by removing two transform nets.  Furthermore, our Sampled PointWise$_f$ achieves at least $100\times$ speedup over PointNet++ \cite{qi2017pointnet++}, DGCNN \cite{wang2018dynamic} and SO-Net \cite{li2018so} at the cost of $2\% - 4\%$ performance drop.

	\begin{comment}
	From another point of view, the basic architecture of our method is derived from PointNet \cite{qi2017pointnet} and even more simple by removing two transform nets. However, after finetuning, our Sampled PointWise$_f$ achieves almost the same accuracy on ModelNet40 as PointNet \cite{qi2017pointnet} while the inference time is $32\times$ more efficient than PointNet \cite{qi2017pointnet}.
	\end{comment}

	\begin{comment}
	From another point of view, our method is derived from PointNet\cite{qi2017pointnet} and even our nework is more simple by removing two transform nets. However, our method achieves almost the same accuracy on ModelNet as PointNet\cite{qi2017pointnet} while our method's feature extracting time is $32\times$ more efficient than PointNet\cite{qi2017pointnet} on Intel i7-8700 CPU(single-core mode).
	\end{comment}
	\subsection{3D Object Retrieval} \label{retrieval}
		\begin{table*}
		\centering
		\small
			\caption{\label{tbl_shrec17_metric_table} Results and best competing methods for the SHREC17 competition. These metrics are computed under micro context.}
			\vspace{4pt}
		\begin{tabular}{p{4cm}<{\centering}|p{2.5cm}<{\centering}|p{1.5cm}<{\centering}p{1.5cm}<{\centering}p{1.5cm}<{\centering}p{1.5cm}<{\centering}p{1.5cm}<{\centering}}           
			\hline
			Method      &  Representation    & P@N & R@N   & F1@N  & mAP   & NDCG  \\ \hline
			
			RotationNet, \cite{kanezaki2018rotationnet} & multi 2D views & \textbf{0.810} & 0.801 & \textbf{0.798} & \textbf{0.772} & \textbf{0.865} \\
			Improved GIFT , \cite{savva2016shrec} & multi 2D views & 0.786 & 0.773 & 0.767 & 0.722 & 0.827  \\
			MVCNN , \cite{su2015multi} & multi 2D views & 0.770 & 0.770 & 0.764 & 0.735 & 0.815  \\
			REVGG, \cite{savva2016shrec} & multi 2D views & 0.765 & \textbf{0.803} & 0.772 & 0.749 & 0.828  \\
			
			\hline
			PointWise & points& 0.747 & 0.746 & 0.740 & 0.708 & 0.808   \\
			\hline
			Ours (Sampled PointWise)& points& 0.728 & 0.731 & 0.721 & 0.690 & 0.788  \\
			\hline
			Ours (Sampled PointWise$_f$)& points& 0.766 & 0.761 & 0.755 & 0.726 & 0.820   \\
			\hline
		\end{tabular}
		\vspace*{-10pt}
	\end{table*}
\begin{table}[t!]
	\small
	\centering
	\caption{Memory demand under different numbers of voxels. The metric is tested on ModelNet40 using Sampled PointWise$_f$.}
	\vspace{6pt}
	\begin{tabular}[width=\linewidth]{c|c|c|c}
		\hline
		Number of Voxels          &Memory Size     & accuracy & accuracy \\
		~& (MB)&avg. class & overall \\ \hline
		$25^3$          &  15 &85.87 & 88.86 \\ \hline
		$50^3$           &  122 &86.40& 89.49  \\ \hline
		$100^3$           &  976 &86.23& 89.35  \\ \hline
		$200^3$           &  8000 &86.43& 89.51  \\ \hline
	\end{tabular}
	
	\label{tab:memory}
\end{table}

	For 3D object retrieval tasks, we stack metric learning model $M$ (a multi-layer perception  with 3 layers) on $\mathcal{F}$ with joint supervision of triplet loss
	and softmax loss. We use the output of the layer before the score
	prediction layer as our feature vector. We conduct the experiments under three large-scale 3D shape benchmarks, including ModelNet40, ModelNet10 and ShapeNet-Core55 \cite{savva2016shrec}.
	
	\textbf{ModelNet}. In Table \ref{tbl_modelnet_retrieval_table}, we compare our method with other state-of-the-art methods\footnote{http://modelnet.cs.princeton.edu/}  in terms of mean average precision (mAP) on ModelNet10 and ModelNet40 benchmarks.
	
	Apart from our methods, other methods  are based on multi 2D views. In addition, PVNet \cite{you2018pvnet} even  integrates
	both the point cloud and the multi-view data  as the input. However, we demonstrate that our Sampled PointWise$_f$ can still achieve comparable performance on 3D object retrieval tasks although we only take point cloud as our input while gaining a remarkable lead in speed. There is still a small gap between our method and other multi-view based methods, which we think is due to the loss of fine geometry details that can be captured by rendered images.
		\begin{figure}[t!]
		\centering
		\small
		\subfigure[]{\label{fig_potential_field}\includegraphics[width=0.23\textwidth,height = 0.1555\textheight]{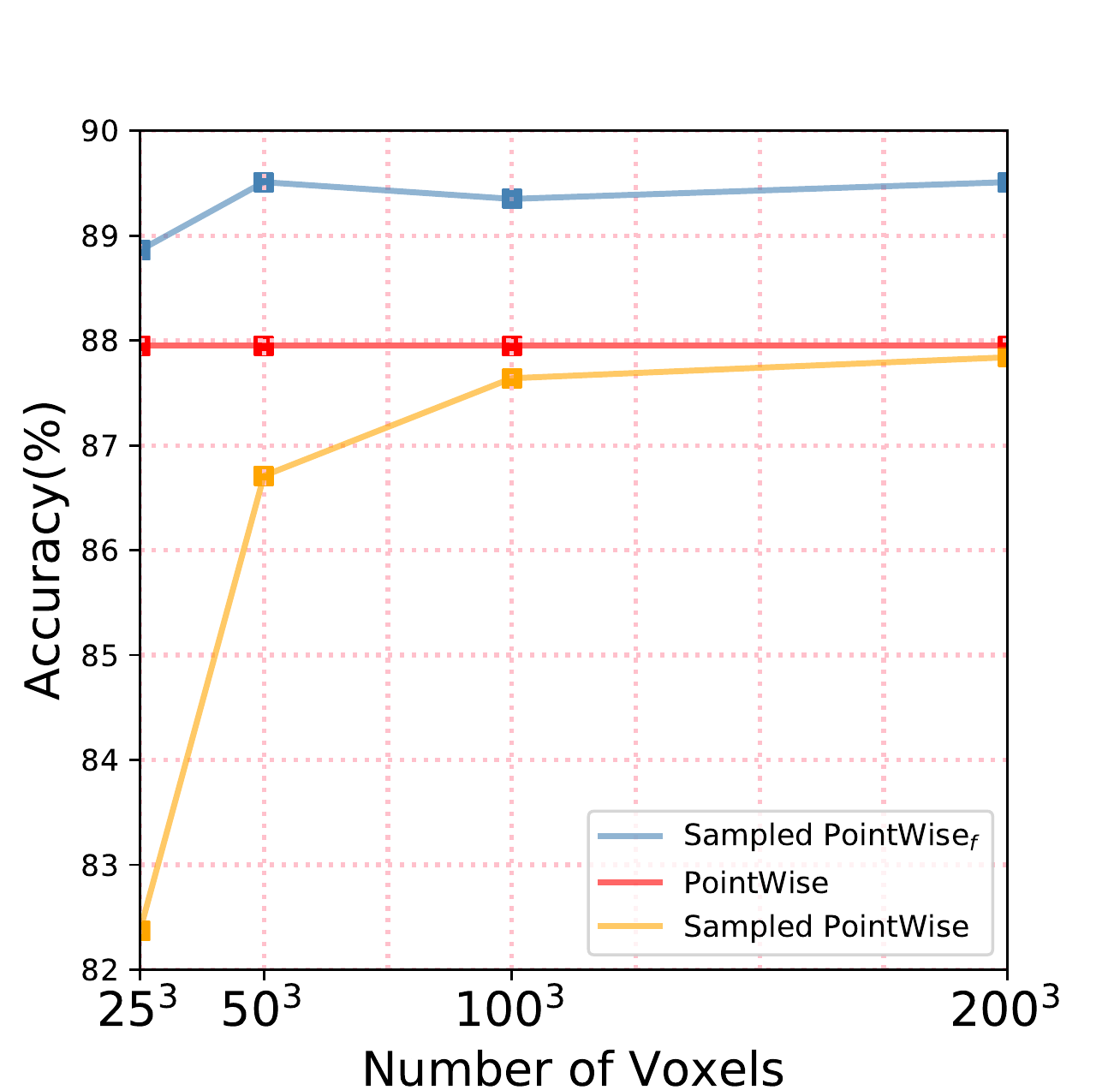}}
		\subfigure[]{\label{fig_som_result}\includegraphics[width=0.23\textwidth,height = 0.1555\textheight]{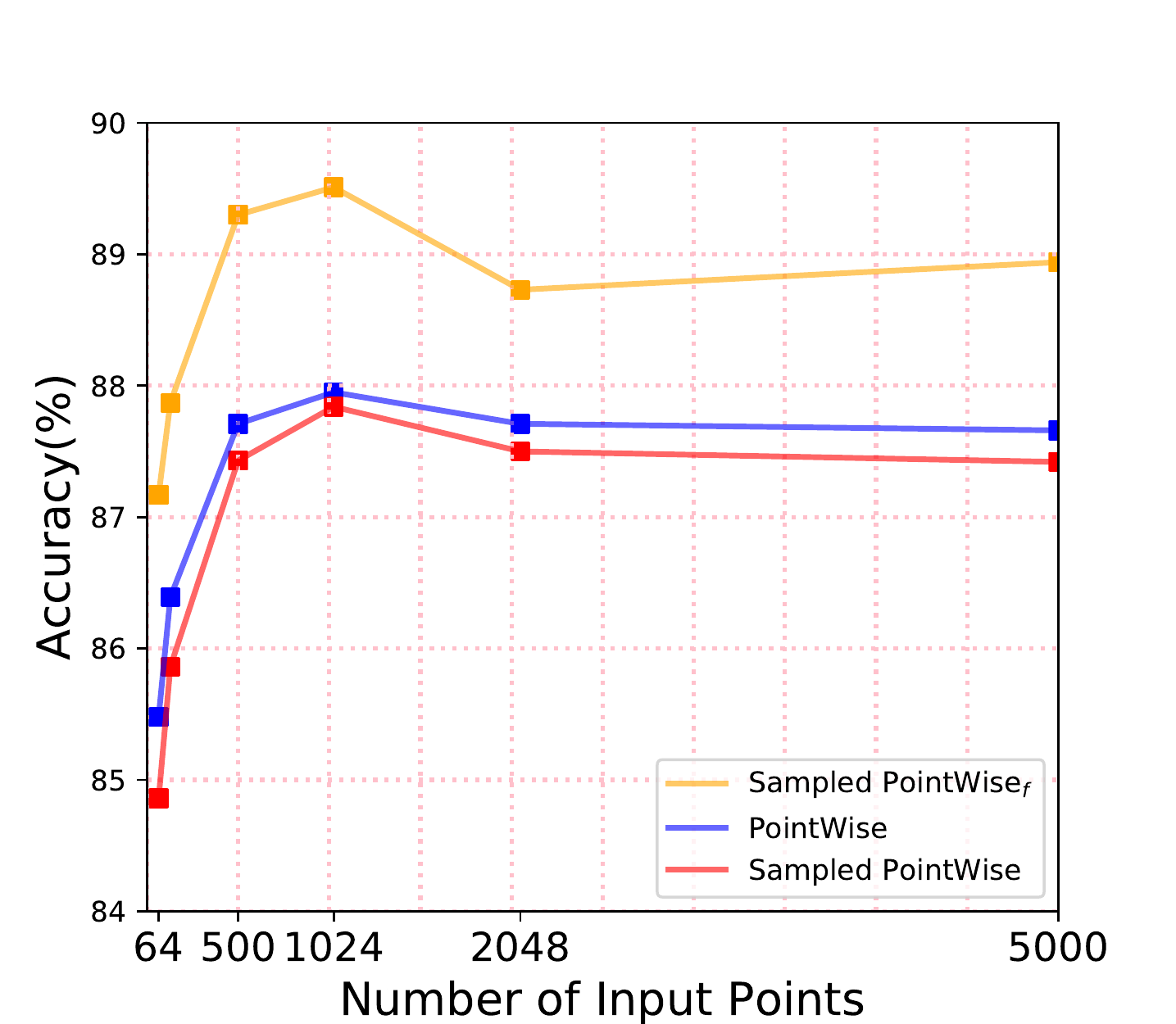}}
		\caption{(a) Effects of the number of voxels  (b) Effects of the number of input points. 
			The metric is overall classification accuracy on ModelNet40.}
		\label{tab:effects}
		\vspace{-6pt}
	\end{figure}
	
	\begin{comment}
	\noindent \textbf{ShapeNet}\ SHREC17\cite{savva2016shrec} provides several evaluation metrics including
	Precision-Recall curve, F-score, mAP, normalized discounted cumulative gain (NDCG).
	These metrics are computed under two contexts - macro and
	micro. Macro metric is a simple average across all categories
	while micro metric is a weighted average according to the number of shapes in each category. For each shape in the test set, we calculate the L2 feature distance with shapes in same predicted category. In Table \ref{tbl_shrec17_metric_table}, we show the retrieval results of our method compared to other state-of-the-art methods on SHREC17\cite{savva2016shrec}.
	We have demonstrated that our method can maintain impressive
	performance under three large-scale 3D shape benchmarks for 3D object retrieval.
	\end{comment}
	 \textbf{ShapeNet}. SHREC17 \cite{savva2016shrec} provides several evaluation metrics including
	Precision, Recall, F1, mAP, normalized discounted cumulative gain (NDCG). We evaluate our methods using the official metrics and compare to the top four
	competitors.

	As shown in Table \ref{tbl_shrec17_metric_table}, our Sampled PointWise$_f$  is about 4\%  worse than RotationNet \cite{kanezaki2018rotationnet} (best perfomance) but only about 1\% worse than Improved GIFT \cite{savva2016shrec}, MVCNN \cite{su2015multi} and REVGG \cite{savva2016shrec} in terms of F1, mAP, NDCG metrics. The main four competitors use multi-2D views as input representations and their network architectures are complicated that are highly specialized to the SHREC17 \cite{savva2016shrec} task.  Given the rather simple architecture of our model and the lossy input representation we use(only point cloud), we interpret our performance as strong empirical support for the effectiveness of our method.  More importantly, our method is extremely fast with only a fraction of computational complexity.
	\subsection{Effects of Number of Voxels}
	Here we show how our model’s performance changes on ModelNet40 for  3D object classification with regard to the number of voxels. As shown in the Fig.~\ref{tab:effects} (a), we can see that
	performance grows as we increase the number of voxels. When $S = 200$, the accuracy of Sampled PointWise is only 0.1\% worse than PointWise. Moreover, after we fine-tune the model $M$ (classification module), no matter what the number of voxels is, the performance of Sampled PointWise$_f$ can achieve almost the same accuracy as PointWise and even $1\%-2\%$  better than PointWise. It means that we can use less voxels, i.e. less memory demand on condition of maintaining comparable performance. In our experiments, it turns out that even with $S = 25 $, our Sampled PointWise$_f$ can achieve 88.86\% recognition rate  with only 15MB memory occupied while the performance merely drops 0.4\% compared with PointNet \cite{qi2017pointnet}. In Table \ref{tab:memory}, we use 8 bits to encode the function value as we have explained in Sec.~\ref{method} and summarize the comparisons of memory demand under different numbers of voxels.
	\begin{comment}
	
	\begin{table}[t!]
	\centering
	\begin{tabular}[width=4.0\textwidth]{c|c|c}
	\hline
	Sampling Interval      & Memory & Accuracy\\  \hline
	0.08          &  15MB & 88.86 \\ \hline
	0.04          &  122MB & 89.51 \\ \hline
	0.02          &  976MB & 89.35  \\ \hline
	0.01          &  8000MB & 89.51  \\ \hline
	\end{tabular}
	\caption{Effects of sampling interval in memory demand. The metric is overall classification accuracy on ModelNet40}
	\label{tab:memory}
	\end{table}
	\end{comment}

	\subsection{Effects of Number of Input Points}
	To evaluate the importance of number of input points, we compare our method on ModelNet40 for classification by using different numbers of points as input. In Fig.~\ref{tab:effects} (b), we can find that
	performance grows as we increase the number of input points
	however it saturates at around 1024 points.
	
	Meanwhile, we fine-tune the classification module and demonstrate that  we can achieve the same accuracy as the PointNet \cite{qi2017pointnet} by only using 500 points. More surprisingly even with 64 points as
	input, our method can still show comparable performance, only 2\% worse than PointNet \cite{qi2017pointnet}.

	\begin{comment}

	Due to the fact that our method’s  time is linear in the number of input points,it indicate that our method can be faster by only using 500 and even 64 points.
	To better illustrate the speed advantage,we give
	a deeper analysis of the time cost under different numbers of input points in Table5.As the Table5 shows, our method make it possible to control the average query time very quick and even within one second by using 64 points as input.
	\end{comment}
	\begin{comment}
	\begin{table}[h]
	\centering
	\small
	\begin{tabular}[width=\linewidth]{c|c|c}
	\hline
	Number of Points & Speed(ms) & Accuracy\\  \hline
	64          &  1.11 & 87.17 \\ \hline
	128          & 1.58  & 87.87 \\ \hline
	500          & 4.30  & 89.30  \\ \hline
	1024          &  7.95 & 89.51  \\ \hline
	\end{tabular}
	\caption{\textbf{Effects of number of points in speed.} The metric is overall classification accuracy on ModelNet40
	test set }
	\label{tab:numberofpoints}
	\end{table}
	\end{comment}

	\section{Limitation}
	Currently, our method requires that the deep architecture must be point-wise. It is not applicable for                                                                                                                                   recently proposed deep architectures such as DGCNN \cite{wang2018dynamic} and SO-Net \cite{li2018so}. Considering its simplicity and speedup performance, how to mine more local information from the point-wise architecture remains to be an attractive and challenging problem. We leave this problem for the future work.
	\begin{comment}
		 Anther interesting phenomenon in our experiments is that we can get relatively high performance for a very coarse  lookup table. This may suggest the possibility of using polynomial function and optimize the overall loss with large batch, thus reducing the learning time and improving model performance.
	\end{comment}

	\section{Conclusion}
	In this paper, we propose to apply a classical lookup table to speed up the inference process for a particular deep architecture for 3D point cloud tasks. This architecture implements a deep function from a set of 3-variable functions by max pooling operation. Our method can ensure the inference time be determined by the memory access no matter how complicated the deep function is. The experiments show that we can achieve almost the same performance with only 15MB memory while gaining $32\times$ speedup over the original PointNet \cite{qi2017pointnet} architecture during the inference process.
	\bibliographystyle{IEEEbib}
	\bibliography{icme2019template}

\begin{thebibliography}{10}

\bibitem{qi2017pointnet}
Charles~R Qi, Hao Su, Kaichun Mo, and Leonidas~J Guibas,
\newblock ``Pointnet: Deep learning on point sets for 3d classification and
  segmentation,''
\newblock {\em Proc. Computer Vision and Pattern Recognition (CVPR), IEEE},
  vol. 1, no. 2, pp. 4, 2017.

\bibitem{qi2017pointnet++}
Charles~Ruizhongtai Qi, Li~Yi, Hao Su, and Leonidas~J Guibas,
\newblock ``Pointnet++: Deep hierarchical feature learning on point sets in a
  metric space,''
\newblock in {\em Advances in Neural Information Processing Systems}, 2017, pp.
  5099--5108.

\bibitem{wang2018dynamic}
Yue Wang, Yongbin Sun, Ziwei Liu, Sanjay~E Sarma, Michael~M Bronstein, and
  Justin~M Solomon,
\newblock ``Dynamic graph cnn for learning on point clouds,''
\newblock {\em arXiv preprint arXiv:1801.07829}, 2018.

\bibitem{li2018so}
Jiaxin Li, Ben~M Chen, and Gim~Hee Lee,
\newblock ``So-net: Self-organizing network for point cloud analysis,''
\newblock in {\em Proceedings of the IEEE Conference on Computer Vision and
  Pattern Recognition}, 2018, pp. 9397--9406.

\bibitem{han2015deep}
Song Han, Huizi Mao, and William~J Dally,
\newblock ``Deep compression: Compressing deep neural networks with pruning,
  trained quantization and huffman coding,''
\newblock {\em arXiv preprint arXiv:1510.00149}, 2015.

\bibitem{denton2014exploiting}
Emily~L Denton, Wojciech Zaremba, Joan Bruna, Yann LeCun, and Rob Fergus,
\newblock ``Exploiting linear structure within convolutional networks for
  efficient evaluation,''
\newblock in {\em Advances in neural information processing systems}, 2014, pp.
  1269--1277.

\bibitem{ba2014deep}
Jimmy Ba and Rich Caruana,
\newblock ``Do deep nets really need to be deep?,''
\newblock in {\em Advances in neural information processing systems}, 2014, pp.
  2654--2662.

\bibitem{sfikas2017exploiting}
Konstantinos Sfikas, Theoharis Theoharis, and Ioannis Pratikakis,
\newblock ``Exploiting the panorama representation for convolutional neural
  network classification and retrieval,''
\newblock in {\em Eurographics Workshop on 3D Object Retrieval}. The
  Eurographics Association, 2017, vol.~8.

\bibitem{shi2015deeppano}
Baoguang Shi, Song Bai, Zhichao Zhou, and Xiang Bai,
\newblock ``Deeppano: Deep panoramic representation for 3-d shape
  recognition,''
\newblock {\em IEEE Signal Processing Letters}, vol. 22, no. 12, pp.
  2339--2343, 2015.

\bibitem{you2018pvnet}
Haoxuan You, Yifan Feng, Rongrong Ji, and Yue Gao,
\newblock ``Pvnet: A joint convolutional network of point cloud and multi-view
  for 3d shape recognition,''
\newblock in {\em 2018 ACM Multimedia Conference on Multimedia Conference}.
  ACM, 2018, pp. 1310--1318.

\bibitem{han2019seqviews2seqlabels}
Zhizhong Han, Mingyang Shang, Zhenbao Liu, Chi-Man Vong, Yu-Shen Liu, Matthias
  Zwicker, Junwei Han, and CL~Philip Chen,
\newblock ``Seqviews2seqlabels: Learning 3d global features via aggregating
  sequential views by rnn with attention,''
\newblock {\em IEEE Transactions on Image Processing}, vol. 28, no. 2, pp.
  658--672, 2019.

\bibitem{sfikas2018ensemble}
Konstantinos Sfikas, Ioannis Pratikakis, and Theoharis Theoharis,
\newblock ``Ensemble of panorama-based convolutional neural networks for 3d
  model classification and retrieval,''
\newblock {\em Computers \& Graphics}, vol. 71, pp. 208--218, 2018.

\bibitem{bai2016gift}
Song Bai, Xiang Bai, Zhichao Zhou, Zhaoxiang Zhang, and Longin Jan~Latecki,
\newblock ``Gift: A real-time and scalable 3d shape search engine,''
\newblock in {\em Proceedings of the IEEE Conference on Computer Vision and
  Pattern Recognition}, 2016, pp. 5023--5032.

\bibitem{wu20153d}
Zhirong Wu, Shuran Song, Aditya Khosla, Fisher Yu, Linguang Zhang, Xiaoou Tang,
  and Jianxiong Xiao,
\newblock ``3d shapenets: A deep representation for volumetric shapes,''
\newblock in {\em Proceedings of the IEEE conference on computer vision and
  pattern recognition}, 2015, pp. 1912--1920.

\bibitem{savva2016shrec}
Manolis Savva, Fisher Yu, Hao Su, M~Aono, B~Chen, D~Cohen-Or, W~Deng, Hang Su,
  Song Bai, Xiang Bai, et~al.,
\newblock ``Shrec’16 track large-scale 3d shape retrieval from shapenet
  core55,''
\newblock in {\em Proceedings of the eurographics workshop on 3D object
  retrieval}, 2016.

\bibitem{kanezaki2018rotationnet}
Asako Kanezaki, Yasuyuki Matsushita, and Yoshifumi Nishida,
\newblock ``Rotationnet: Joint object categorization and pose estimation using
  multiviews from unsupervised viewpoints,''
\newblock in {\em Proceedings of IEEE International Conference on Computer
  Vision and Pattern Recognition (CVPR)}, 2018.

\bibitem{su2015multi}
Hang Su, Subhransu Maji, Evangelos Kalogerakis, and Erik Learned-Miller,
\newblock ``Multi-view convolutional neural networks for 3d shape
  recognition,''
\newblock in {\em Proceedings of the IEEE international conference on computer
  vision}, 2015, pp. 945--953.

\end{thebibliography}
	
\end{document}